\documentclass{article}

\usepackage{arxiv}
\usepackage{graphicx}
\usepackage{comment}
\usepackage{amsmath,amssymb} % define this before the line numbering.
\usepackage{color}

\usepackage[utf8]{inputenc} % allow utf-8 input
\usepackage[T1]{fontenc}    % use 8-bit T1 fonts
\usepackage{url}            % simple URL typesetting
\usepackage{booktabs}       % professional-quality tables
\usepackage{amsfonts}       % blackboard math symbols
\usepackage{nicefrac}       % compact symbols for 1/2, etc.
\usepackage{microtype}      % microtypography
\usepackage{lipsum}

\usepackage{subcaption}
\usepackage{algorithm}
\usepackage{algorithmic}
\usepackage{xcolor}
\usepackage{bm}
\usepackage[pagebackref=true,breaklinks=true,letterpaper=true,colorlinks,bookmarks=false]{hyperref}

\usepackage{algorithm}
\usepackage{collcell}
\usepackage{amssymb} %checkmark

\usepackage{xcolor}
\definecolor{darkergreen}{RGB}{21, 152, 56}
\definecolor{red2}{RGB}{252, 54, 65}

\title{Self-Supervised Learning Using Consistency Regularization of Spatio-Temporal Data Augmentation for Action Recognition}

\author{
  Jinpeng Wang \\
  School of Electronics and Information Technology\\
  Sun Yat-sen University\\
  Guangzhou, 510006, China \\
  \texttt{wangjp23@mail2.sysu.edu.cn} \\
   \And
  Yiqi Lin \\
  School of Data and Computer Science\\
  Sun Yat-sen University\\
  Guangzhou, 510006, China \\
  \texttt{linyq29@mail2.sysu.edu.cn} \\
  \AND
  Andy J. Ma \thanks{Corresponding author.} \\
  School of Data and Computer Science\\
  Sun Yat-sen University\\
  Guangzhou, 510006, China \\
  \texttt{majh8@mail.sysu.edu.cn} \\
}

\begin{document}
\maketitle

\begin{abstract}
Self-supervised learning has shown great potentials in improving the deep learning model in an unsupervised manner by constructing surrogate supervision signals directly from the unlabeled data. Different from existing works, we present a novel way to obtain the surrogate supervision signal based on high-level feature maps under consistency regularization. In this paper, we propose a Spatio-Temporal Consistency Regularization between different output features generated from a siamese network including a clean path fed with original video and a noise path fed with the corresponding augmented video. Based on the Spatio-Temporal characteristics of video, we develop two video-based data augmentation methods, i.e., Spatio-Temporal Transformation and Intra-Video Mixup. Consistency of the former one is proposed to model transformation consistency of features, while the latter one aims at retaining spatial invariance to extract action-related features. Extensive experiments demonstrate that our method achieves substantial improvements compared with state-of-the-art self-supervised learning methods for action recognition. When using our method as an additional regularization term and combine with current surrogate supervision signals, we achieve 22\% relative improvement over the previous state-of-the-art on HMDB51 and 7\% on UCF101.
\end{abstract}

% keywords can be removed

%suppressing static features but 
\section{Introduction}

%============================ first paragraph==========================
%history and background
%======================================================================
Convolutional neural networks (CNNs) can achieve competitive accuracy on a variety of video understanding tasks, including action recognition \cite{hara2018can}, temporal action detection \cite{zhao2017temporal} and spatio-temporal action localization \cite{weinzaepfel2015learning}.
Such success is built on the heavily annotated datasets, which are time-consuming and expensive to obtain.
Since numerous unlabeled data is instantly available (e.g. online video sequences), it has drawn more and more attention from the community to utilize the off-the-shelf unlabeled data to improve the performance of CNNs. 

%============================ second paragraph==========================
%related work in self-supervised
%======================================================================
\begin{figure}[t]
	\centering
	\includegraphics[width=.8\linewidth]{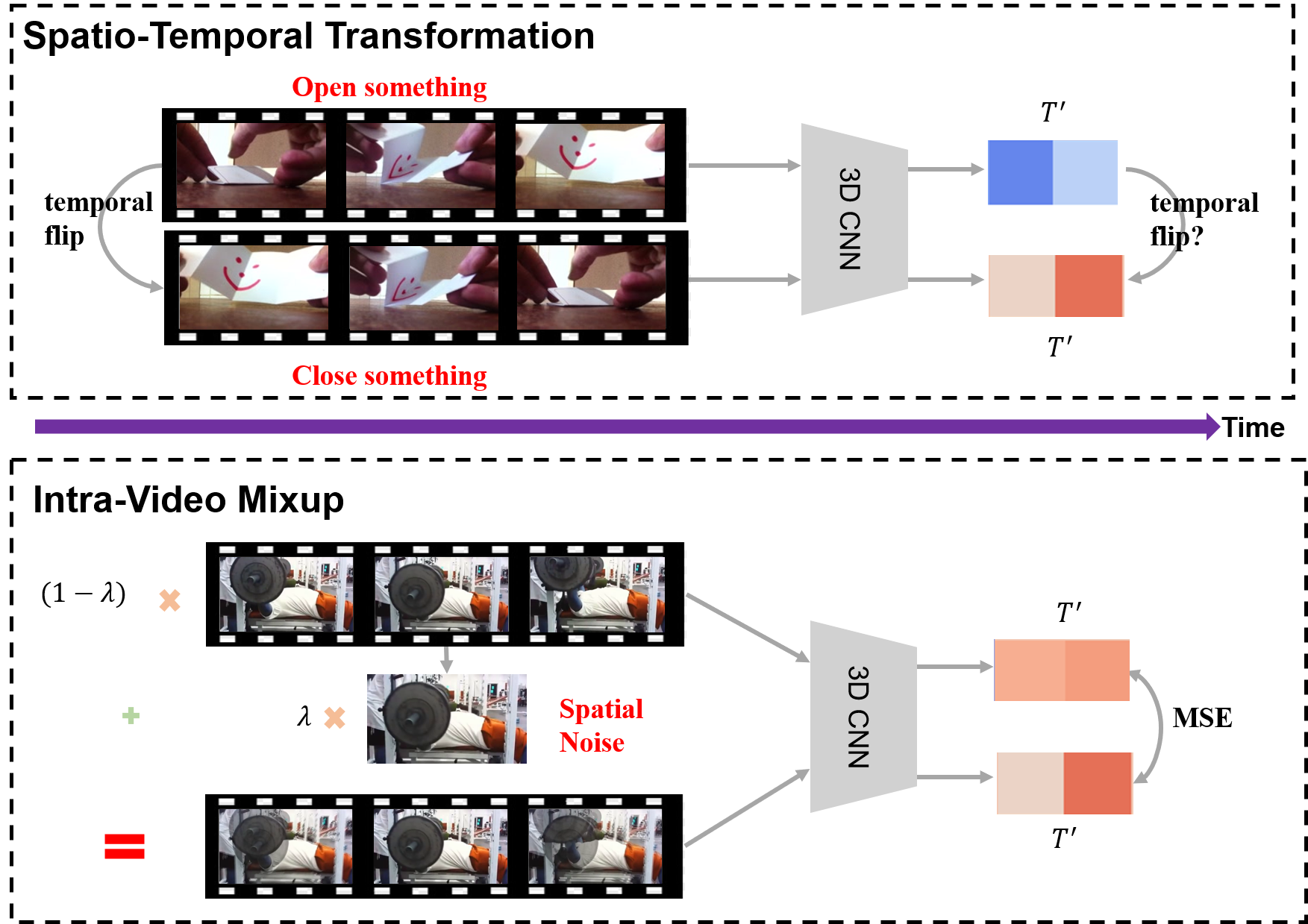}
	\caption{
	Top: We enforce 3D CNN keep transformation consistency at high-level feature map.
	%Top: A good 3D CNN should keep transformation consistency in high-level feature map.
	Bottom: Action category is mainly describe by motion pattern, the video keep semantic consistency by introducing fixed noise frame-wise. Here
    $\lambda$ is the probability of interpolation and $T'$ is the length of time dimension.
    }
	\label{fig:TC_bk}
\end{figure}

%	Top: The semantic understanding of video sequences highly depend on the temporal order, the video via Temporal Flip may have completely opposite semantics. Bottom: Action category is mainly describe by motion pattern, the video keep consistency by introducing fixed noise frame-wise.

Self-supervised learning aims at resolving the designed \emph{pretext} tasks which can be formulated using only unlabeled data \cite{zhai2019s}. 
Without huge human efforts for annotation, surrogate supervision signals can be used to pretrain the CNNs with the pretext tasks. 
For instance, the relative positions of image patches \cite{noroozi2016unsupervised} and image rotation degrees \cite{gidaris2018unsupervised} are used as surrogate supervision signals to learn semantic features for better image classification. 

Besides image applications, self-supervised learning has been extended to video applications especially action recognition.
Using temporal information in videos, recent works attempt to perform temporal modeling among frames via specifically designed supervision signals, such as the arrow of time \cite{wei2018learning} or the clip order \cite{xu2019self}. 
However, current video datasets usually exist large \emph{implicit biases} over scene and object structure, which makes the temporal structure become less insignificant. 
Consequently, these methods may focus on appearance information rather than useful video representation.
%However, these methods may not be robust to large \emph{implicit biases} of scene and object structures.

% The most widely used video datasets are UCF101 \cite{soomro2012ucf101} and HMDB51 \cite{kuehne2013hmdb51}.
% However, these datasets have two obvious disadvantages: \emph{i}. The size of them is very small (only thousand of samples). \emph{ii}. These datasets exist large \emph{implicit biases} over scene and object structure, which makes the temporal structure become less insignificant \cite{girdhar2020cater}.
% Under the influence of these disadvantages, a conventional self-supervised learning method is easy to overfitting and may focus on static image feature rather than general semantic feature. 
% In this work, we focus on ease this problem via well-designed data-augmentation.

%In this work, we given a new aspect: start from the data itself and let the model pay more attention to the temporal information rather than implicit biases by data-augmentation.

%============================ third paragraph==========================
%in this work we propose new method
%======================================================================

To overcome this limitation, we propose a novel surrogate supervision signal based on \emph{consistency regularization} of spatially and temporally augmented data.
In the proposed method, data augmentation is applied to the original input. Then the original and augmented videos are fed into the simease CNN which is trained to minimize the mean-square error (MSE) between the corresponding features of the original and augmented videos.
By this way, a new pretext task can be defined for pretraining using the augmented data.
As a result, data augmentation becomes a very important component in the proposed pretext task to extract discriminative features for action recognition.
Based on the Spatio-Temporal characteristics of video, this paper proposes Spatio-Temporal Transformation and Intra-Video Mixup as shown in Figure \ref{fig:TC_bk} for data augmentation. 
By minimizing the MSE between the features of the original and augmented videos, a CNN model can be pretrained semantically in two aspects, i.e., robustness to transformation and spatial invariance. 

%1) the transformation on input data should also keep consistency on feature maps through the CNN. Second, it will learn to suppress static image feature and keep spatial invariance. Throughout the work we support these arguments quantitatively.

%============================ fourth paragraph==========================
%fourth:our work performance 
%======================================================================

Experimental results show that the proposed self-supervised learning approach using consistency regularization provides a powerful surrogate supervision signal for semantic feature learning and leads to dramatic improvements on action recognition benchmarks.
By incorporating the proposed consistency regularization into existing self-supervised learning methods, performance can be further improved.  
The major contributions of this paper are summarized as follows:
\begin{itemize}
\item We propose a novel consistency regularization of spatially and temporally augmented data, which can be served as a new supervised signal under the self-supervised framework and can easily improve existing methods for action recognition.
\item We propose two novel data augmentation methods, Spatio-Temporal Transformation and Intra-Video Mixup, for transformation consistency and spatial invariance in video-based self-supervised learning.
\item Our proposed method outperforms state-of-the-art methods on UCF101 \cite{soomro2012ucf101} and HMDB51 \cite{kuehne2013hmdb51}.
\end{itemize}
 
\section{Related Work}
% This work relates to two topics in machine learning and computer vision: Self-supervised Learning, Consistency Regularization and Data Augmentation.

\subsection{Self-supervised Learning}
Self-supervised learning is a generic learning framework that relies on proxy tasks that can be formalized using only unsupervised data.
Gidaris et al.
\cite{gidaris2018unsupervised} use the prediction of image rotation as training target while we just use Spatial Rotation as perturbation to make model more robust.
Meng et al. \cite{ye2019unsupervised} adopt simease network and adopt image instance classification but we focus on video domain.
More recently, this valuable research has expanded into video domain. The related work \cite{xu2019self} predicts video clip order or learning the arrow of time \cite{wei2018learning,kim2019self}. 
However, these methods are specially designed without generalization, which is our focus.

\subsection{Consistency Regularization}

Consistency regularization applies data augmentation to semi-supervised learning, which takes advantage of the idea that a classifier should output the same class distribution for an unlabeled example even it has been augmented. 
There are many semi-supervised learning based on consistency regularization in general \cite{sajjadi2016regularization,tarvainen2017mean,berthelot2019mixmatch,xie2019unsupervised}.
In this work, we propose a novel form of consistency regularization through the use of Spatio-temporal Transformation and Intra-Video Mixup. To our best knowledge, we are the first one that introduce consistency regularization into self-supervised learning. 

Our motivation is also different from \cite{Guo_2019_CVPR} which involves geometric transformation in attention consistency under fully supervised setting. Note that, keeping visual attention consistency relies on the class activation map (CAM) \cite{zhou2016learning} which needs category annotations.
While, we restrain high-level feature map without using CAM. Otherwise, we add constraint along the temporal dimension rather than pixel-wise in spatial.

\subsection{Data Augmentation}

In image classification, the input image will usually undergo elastic deformation or add noise, which can greatly change the pixel content of the image without changing the label. Based on this, a lot of augmentation technologies include rotation, flip and color jittering \cite{krizhevsky2012imagenet} has been proposed.  Recently, Mixup \cite{zhang2018mixup} is a practical data augmentation method in image classification that combines two samples, the ground-truth label of the new sample is given by the linear interpolation of one-hot labels. 

However, Mixup requires both label and pair samples in a batch during train. In contrast, Intra-Video Mixup selects one frame from the video itself and needn't label under the framework of consistency learning, which can be integrated into unsupervised or semi-supervised learning.

\section{Methodology}

In this section we present the proposed Spatio-Temporal Consistency Regularization (STCR) for action recognition. 
We first give an overall description of STCR in Section \ref{sec:overall_archite}. Then, details about Spatio-Temporal Data Augmentation and Consistency Regularization are introduced in Section \ref{sec:Spatio-TemporalAugmentation} and Section \ref{sec:TemporalConsistencyRegularization}. At last, we discuss the advantages of STCR in Section \ref{sec:discussion}.
%We first propose our new regularization technique, which solving it in a self-supervised manner requires learning of a useful video representation. Afterwards, we transfer the learned encoder $F$ to downstream action recognition task.
%Our propose is to learn a strong encoder $F$ with our regularization technique in a self-supervised manner, which can be further transfer to downstream action recognition task.
%We first propose our new regularization technique, which solving it requires learning of a useful video representation. Afterwards, we introduce .

%The training data consist of total of $N$ inputs and each input clip is denoted ${x}$, where ${x} \in \mathbb{R}^{C \times T \times H \times W}$. $C$, $T$, $H$ and $W$ denotes the number of channel, the number of frames, the height of single frame and the width of single frame, respectively.

\subsection{Overall Architecture}
\label{sec:overall_archite}

\begin{figure*}[hbt!]
	\centering
	\includegraphics[width=\linewidth]{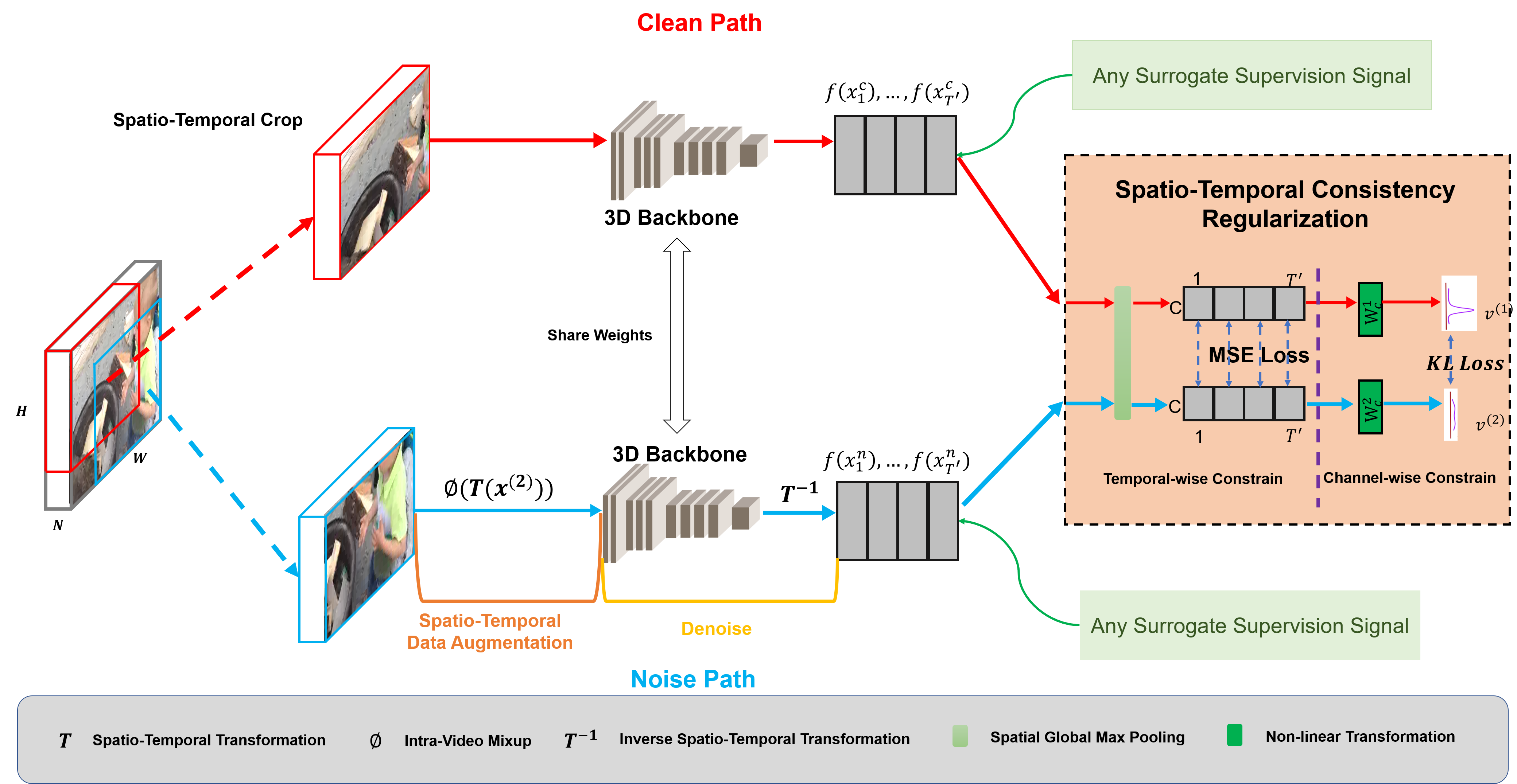}
	\caption{ Framework of the proposed method for training.
	% Diagram of the self-supervised train procedure with Siamese network. 
	The input videos are projected into high-level feature maps with the CNN backbone. 
	The augmented video is constrained to be consistent with the original video in both temporal direction and channel. (Best viewed in color).}
	\label{fig:overall}
\end{figure*}

%Stochastic data augmentation is applied to an unlabeled video, and each augmented video is fed through the backbone. Then, the temporal consistency loss is calculated. 

%In order to learn semantic feature under an self-supervised manner, we propose Spatio-Temporal Consistency Regularization (STCR) using Spatio-Temporal Transformation and Intra-Video Mixup.

The framework of the proposed self-supervised learning method is shown in Figure \ref{fig:overall}. 
The trunk of our framework is a siamese network with two branches (named as \textit{Clean Path} and \textit{Noise Path}).
Each branch contains a conventional 3D backbone, and the parameters of the backbone between all branches are shared and recorded as $\theta$. 
For each input $x$, these two branches randomly crop a fixed-length clip of video frames from different spatial locations, denoted as ${x}^{c}$ and ${x}^{n}$.
In this way, the input frames between these two branches get different distribution in pixel level but consistent in semantic level.

After that, we directly feed the inputs ${x}^{c}$ into the \textit{Clean Path} and we perform Spatio-Temporal Transformation and Intra-Video Mixup on inputs ${x}^{n}$ for \textit{Noise Path}.
The output of these two branches are represented as $f_{x^{c}}, f_{x^{n}} \in \mathbb{R}^{C' \times T' \times H' \times W'}$, formally written as:
\begin{align} 
        f_{ x^{c}} &= F(x^{c};\theta)\\
        f_{ x^{n}} &= F(\phi (T(x^{n})); \theta)
\end{align}
where$F(\cdot ; \theta)$ represents the siamese branch with parameter $\theta$, $T$ represents Spatio-Temporal Transformation and $\phi$ represents Intra-Video Mixup. $C'$ is the number of channel and $T'$ is the length of time dimension, $W'$ and $H'$ is the spatial size. 

At last, these extracted features are fed into the regularization head and pretext task head which can be any usual self-supervised learning setting, such as predicting the arrow of time. 
For the Temporal Consistency Regularization, we minimize the gap of high-level features between the \textit{Clean Path} and \textit{Noise Path} with additional corresponding inverse transformation. The Algorithm \textcolor[rgb]{1,0,0}{1} in Supplementary Section \textcolor[rgb]{1,0,0}{2} summarizes the proposed method.

\subsection{Spatio-Temporal Data Augmentation}
\label{sec:Spatio-TemporalAugmentation}
%\subsection{Spatio-Temporal Transformation}
\subsubsection{Spatio-Temporal Transformation}
\label{sec:SpatialEnhancement}

\begin{figure}[hbt!]
	\centering
	\includegraphics[width=.95\linewidth]{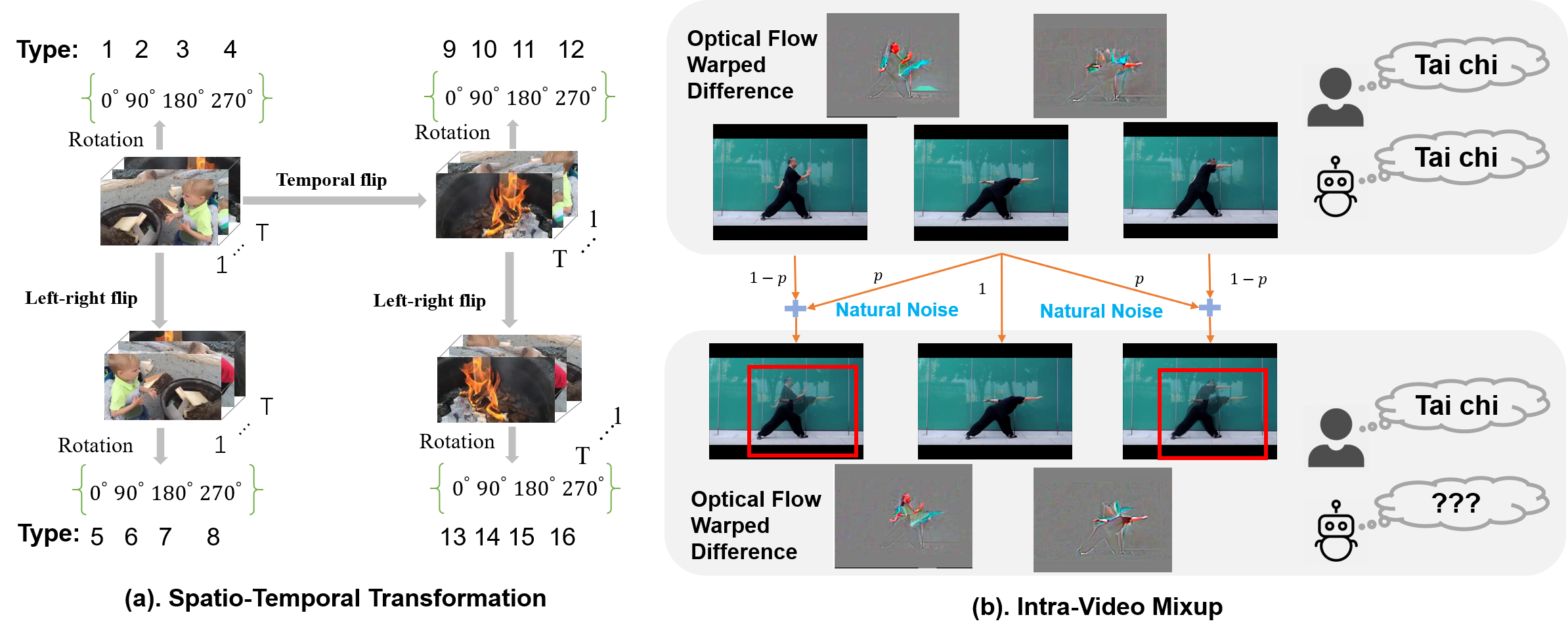}
	\caption{ An illustration of Spatio-Temporal Transformation (STT) and Intra-Video Mixup (IM), STT include four types of flip and each kind of filp corresponding to four types of rotation.
	STT encourages base encoder keep the direction of motion information in high-level feature map.
	IM encourages the model be not sensitive to static spatial appearance and focus more on dynamic motion information.}
	\label{fig:stt_im}
\end{figure}

There are two targets throughout the design of Spatio-Temporal Transformation, the network must able to learn the potential temporal patterns and the learning process are robust to the spatial variations in frames.
Designed with these objectives in mind, the proposed Spatio-Temporal Transformation $T$ is implemented via the \textit{Rotation} or \textit{Flip} on input clips. 
More specifically, for a random input clip $X$, we random choose \textit{Flip} operator $z_1$ from set $F$ \{No Flip, Left-right Flip, Temporal Flip, Temporal Flip + Left-rigth Flip\} and \textit{Rotation} degrees $z_2$ from set $R$ \{0$^{\circ}$, 90$^{\circ}$, 180$^{\circ}$, 270$^{\circ}$ \}.
The total combinations $L$ is $4 \times 4 = 16$, as shown in Figure \ref{fig:stt_im}.
For an input clip ${x}^{n}$ with random combination $(z_1,z_2)$ , formally,
\begin{equation}
    T({x}^{n}) = Rot(Flip({x}^{n}, z_1), z_2)
\end{equation}
where $Rot$ is the rotation operator and $Flip$ is the Flip operator. All these operators process all frames in the same way throughout the entire video clip. 

%\subsection{Intra-Video Mixup}
\subsubsection{Intra-Video Mixup}
\label{sec:SpatialMixup}

Intra-Video Mixup is proposed to learn the desired semantic structure and to avoid the network focus on low-level visual clues which might cause the degradation of performance. 
Hence, Intra-Video Mixup aims at generating new training data by adding static spatial perturbation without changing its motion pattern.

More specifically, for training sample $x^{n}$ we random select one frame $x^{n}_k$ from itself and then combine them, defined as:
\begin{equation}
     \phi (x^{n}_{j})= (1- \lambda) \cdot x^{n}_{j} +  \lambda \cdot x^{n}_{k}, j \in [1,T]
\end{equation}
where $\lambda$ is sampled from the beta distribution Beta($\alpha$,$\alpha$), $j$ is the index of frame and $k$ denote the index of random selected frame. An intuitive example is shown in Figure \ref{fig:stt_im}.

\subsection{Spatio-Temporal Consistency Regularization} 
\label{sec:TemporalConsistencyRegularization}

The design of STCR has two objectives: 
% First, since our purpose is to learn a semantic relevant CNN with given Pretext task which can transfer to downstream task, the backbone CNN should afford mainly modeling capabilities, that is, the STCR must introduce as few learnable parameters as possible. 
First, since our purpose is to learn a semantic relevant CNN with given Pretext task, the STCR should introduce learnable parameters as few as possible.
Second, the network should focus more on temporal aspect but robust to spatial noise.
To achieve these goals, STCR includes two parts: \textit{temporal-wise} and \textit{channel-wise constrains}.

In particular, we introduce inverse transform $F^{-1}$ on $f_{x_n}$ to align the feature from Clean and Noise Path, which simply reverse the operation of Spatio-Temporal Transformation.
Since temporal information play a key role in video representation, we first encourage \textit{Noise Path} and \textit{Clean Path} keep identify \textit{temporal-wise}.
Formally,
\begin{equation}
    \mathcal{L}_{tw} = {||\psi (f_{ x^{c}}) - \psi (T^{-1}(f_{x^{n}}))||}^2
\end{equation}
where $\psi$ is the explicit feature mapping function from high-level feature map with shape $\mathbb{R}^{C'\times T' \times H' \times W'}$ to $\mathbb{R}^{C' \times T'}$ that keep temporal dimension.
As \textit{Noise Path} and \textit{Clean Path} are randomly cropped from different spatial location from same video, we select no-parameter spatial global max pooling layer as $\psi$. In this way, we only force the max response along each time dimension to be similar.

However, although the majority of Clean Path and Noise Path in spatial dimension is identical, there may still exist some special cases: key motion area is preserved in one path but missing in another path.
In addition, keeping consistent under Spatio-Temporal Transformation along each time dimension may be hard to learn at the beginning of training. 
To address these problems, we propose a channel-wise constrain which is more easy to optimize.
This constrain further reduce the temporal dimension via learnable transformation.
In this way, the input video clip is encoded into a descriptor with shape $C'$. And we compute $\mathcal{L}_{cw}$ by 
\begin{equation}
    \mathcal{L}_{cw} = {KL( \mathcal{H} (\psi(f_{ x^{c}}); W_{c}^{1}), \mathcal{H} (\psi(T^{-1}(f_{x^{n}})); W_{c}^{2}))}
\end{equation}
where $\mathcal{H}$ is a non-linear mapping with parameter $W_c$ and $KL$ means Kullback–Leibler divergence. 
The overall loss includes two terms:
\begin{equation}
    \mathcal L = \mathcal L_{tw} + \gamma \mathcal L_{cw}
\end{equation}{}
To combine these terms, we scale the latter by $\gamma$. 

\subsection{Discussion}
\label{sec:discussion}
In summary, the proposed STCR is advantageous in the following aspects.
% The simple formulation of STCR has several key advantages. 
\emph{i}. Spatio-Temporal Transformation is implemented via matrix flip and transpose operations, which costs almost no additional computation. The pseduo codes of Intra-Video Mixup and Spatio-Temporal Transofrmation are presented in Supplementary Section \textcolor[rgb]{1,0,0}{1}. 
\emph{ii}. The intuition behind Spatio-Temporal Transformation attributes to the transformation-inverse invariance, i.e., the high-level feature maps of the transformed video after inverse transformation need to be consistent with the feature maps of the original video.
% the same transformation between two input videos is the same as the one between the high-level feature maps of them. 
Experimental results verify that Spatio-Temporal Transformation can result in clear performance improvement and high-level feature maps become transformation-consistent as illustrated in Section \ref{sec:vis_tt}.
\emph{iii}. For Intra-Video Mixup, consistence of such augmentation can suppress static visual features but preserve temporal motion patterns, which are discriminative for action recognition. Please refer to Section \ref{sec:deeper_intra_sample_mixup} for detailed experimental analysis. 

\section{Experiments}
\subsection{Datasets and Evaluations}
\label{sec:dataset_and_evaluations}

\begin{table}[t]
 \setlength{\belowcaptionskip}{1pt}
\centering
\tabcolsep=0.1cm
\begin{tabular}{@{}cc@{}}
\hline
Method & Top-1 accuarcy (\%) \\ 
\hline
Baseline & 24.2 \\
Gaussian Noise & 25.8 \\
Video Mixup & 23.1 \\
Video CutMix & 24.0 \\
Inter-Video Mixup & 28.0 \\
Intra-Video Mixup & \textbf{30.2} \\

\hline
\end{tabular}
\caption{Comparison Intra-Video Mixup with other methods.}
\label{tab:mix_cmp}
\end{table}

\subsubsection{Datasets.}
In this section we use two well-known datasets for video self-supervised learning, UCF101 and HMDB51.

\textbf{UCF101} is an action recognition dataset of realistic action videos, collected from YouTube, having 13,320 videos from 101 action categories.

\textbf{HMDB51} is collected from various sources and contains 6,849 clips divided into 51 action categories.

%, mostly from movies, and a small proportion from public databases such as the Prelinger archive, YouTube and Google videos. The dataset

\begin{table}[t]
 \setlength{\belowcaptionskip}{1pt}
    \centering
    {
    \begin{tabular}{cccc}
    \hline STT & IM & UCF101(\%) & HMDB51(\%) \\

    \hline
    & & 60.3 & 24.2 \\
    \checkmark & & 63.2 & 27.9 \\
    & \checkmark & 64.3 & 30.3  \\
    \checkmark & \checkmark & \textbf{68.4} & \textbf{32.2} \\
    \hline
    
    \end{tabular}
    }
    \caption{Top-1 accuracy [\%] obtained by individual methods or combination on UCF101 and HMDB51. STT represent for Spatio-Temporal Transformation and IM for Intra-Video Mixup.}
    \label{tab:ablation_cmp}
\end{table}

\subsubsection{Evaluations.}
\label{sec:evaluations}
We use PyTorch \cite{paszke2017automatic} to implement the whole framework. In order to demonstrate the generality of our work, we use C3D \cite{tran2015learning}, R3D \cite{hara2018can}, R(2+1)D \cite{tran2018closer} and I3D \cite{carreira2017quo} as baseline. For each model, the consistency regularization is performed at before the global average pooling layer. We provide complete implementation details of each network in Supplementary Section \textcolor[rgb]{1,0,0}{2}.
Results of the final model obtained in the two intermediate steps. 

\textbf{Step 1: Self-supervised Learning.}
We train and test our self-supervised learning method on split 1 of UCF101. The input clip length is 16 frames and the temporal stride is 4 frames so that the adjacent frames have great visual difference. (The chosen of temporal stride is analysed in Supplementary Section \textcolor[rgb]{1,0,0}{4}.) Specifically, for each clip, we randomly sample 16 frames with interval as 4, and spatially resize the frames as 112 $\times$ 112 pixels.  We start from learning rate of 0.01, and assign a weight decay of 5e-4. The total epochs is 50 and the learning rate goes down to 1/10 every 10 epochs. $\mathcal{H}$ is implement via non-linear ReLU activation followed by FC layer. In all experiments, we set $\gamma$ to 1e-1 and $\alpha$ as 1 default. 

\textbf{Step 2: Fine-tuning the model.}
Once we finish the pretraining stage, we use our learned parameters to initialize the 3D CNNs for action recognition, while the last fully connected layer is initialized randomly. During the finetuning and testing, we follow the same protocol in \cite{xu2019self} to provide a fair comparison. We apply random spatial cropping and flip horizontal to perform data augmentation. We start from a learning rate of 0.05, and assign a weight decay of 5e-4. The total epochs is 45 and the learning rate goes down to 1/10 every 10 epochs.

As our method is generality and can be easily extended to other video understanding task, we also present the experiment of Video Clip Retrieval task in Supplementary Section \textcolor[rgb]{1,0,0}{6}.

\subsection{Ablation Analysis}
In this section, we give detail ablation study on split 1 of UCF101 and HMDB51. For each experiment, we follow the two steps evaluation on Section \ref{sec:dataset_and_evaluations} and we report Top-1 accuracy (\%). All methods use I3D as baseline.
%and Baseline(scratch) in each table means train from random initialization model.

\textbf{Comparison against Variants of Mixup.}
In this experiment, we point out the merit of using different mixup-based methods under configurations: 
(a). Maps identical spatial white Gaussian Noise to each frame (Gaussian Noise).
(b). Mixes two videos by interpolating the image frame-wise (Video Mixup).
(c). Replacing each frame's one random region with a patch from another frame (Video CutMix).
(d). Random select one frame from another video, and mixes each frame with this frame  (Inter-Video Mixup).
(e). Random select one frame from itself, and mixes each frame with this frame (Intra-Video Mixup).

The result is shown in Table \ref{tab:mix_cmp}.
% We make two observations. \emph{i}. 
% we define temporal consistency???
% data augmentation: prevent overfitting and ...
Without available label, Video Mixup and Video CutMix perform worse than the baseline, which demonstrates the importance of keeping semantic consistency. 
Since the Inter- and Intra- Video Mixup generate semantic-preservation videos, we observe that these two methods are better suited for modeling actions than Gaussion Noise. %Also, the Intra-Video Mixup lead to best result, .
We also observe that Intra-Video Mixup leads to 2.2\% relative gain when compared with Inter-Video Mixup. Notice that the only difference between them is the former selects fixed frame in the same video, we conclude that suppress static image feature is essential for video representation when the size of the dataset is limited.
% Video mixup, however, this mixup samples tend to be unnatural and inferior to Spatial Mixup.

\textbf{Ablation Study.}
In this experiment, we evaluate the effectiveness of each module in STCR independently at first, and then we combine them together. 
% At first, we train I3D from scratch as baseline.
% And then we train several baseline models equipped with our Temporal Transformation or Intra-Video Mixup independently. 
The result is reported in Table \ref{tab:ablation_cmp} and we can make two observations. \emph{i.} Each of two modules leads to improved accuracy when using both UCF101 and HMDB51 as benchmarks, which indicates these modules have succeeded in learning temporal abstractions.
% As the only change between these models is the usage of Flip or Spatial Mixup module, we deduce that these modules have succeeded in learning temporal abstractions. 
\emph{ii.} These two modules can be combined to achieve better results (30\% higher than train from scratch in HMDB51).
%design two experiments, one 
More results about atom operation in Spatio-Temporal Transformation are given in Supplementary Section \textcolor[rgb]{1,0,0}{4}.

\textbf{The influence of backbone.}
We use 4 types of alterations with varying backbone to test our method: (a). C3D, (b). R3D, (c). R(2+1)D, (d). I3D. %The four networks, we follow a global max pooling layer followed by a fully-connected layer with softmax on top of it.
The results of this controlled experiment are shown in Table \ref{tab:different_backbone}. We observe that STCR leads to improved accuracy when using all backbones. 
As the only change between these models is the usage of STCR, we deduce that the STCR framework has succeeded in learning semantic feature and our method has good generality. 
We also observe that I3D and R(2+1)D are more competitive than C3D and R3D.

\begin{table}[t]
 \setlength{\belowcaptionskip}{5pt}
    \centering
    {
    \begin{tabular}{ccc}
    \hline
    Method& UCF101(\%) & HMDB51(\%) \\
    \hline
    C3D(random) & 60.4 & 22.4 \\
    C3D + STCR & 66.4& 29.2 \\
    \hline
    R3D(random) & 54.6 & 21.3\\
    R3D + STCR & 61.2 & 30.4\\
    \hline
    R(2+1)D(random) & 56.6 & 21.4 \\
    R(2+1)D + STCR & \textbf{70.5}&31.9\\
    \hline
    I3D(random)&60.3 &24.2\\
    I3D + STCR &68.4&\textbf{32.2}\\
    \hline
    \end{tabular}
    }
    \caption{ Top-1 accuracy [\%] obtained by different backbones equipped with STCR.}
    \label{tab:different_backbone}
\end{table}

\subsection{Experiments on Benchmarks}
In this experiment, we compare with state-of-the-art self-supervised learning methods in the video domain.
To get the final action recognition result for a video, 10 clips are sampled form the video to get clip predictions, and then averaged to obtain the video prediction. 
And the other settings were the same as Section \ref{sec:evaluations}. 
For a fair comparison, all methods use C3D as backbone. 
As our STCR is a general framework and can be suitable for any self-supervised learning method, except only using STCR as a supervisory, we also combine STCR with \cite{misra2016shuffle} and \cite{xu2019self} and serve as regularization.

\begin{table}[t]
    \centering
    {
    \begin{tabular}{ccc}
    \hline
    Method& UCF101(\%) & HMDB51(\%) \\
    \hline
    Shuffle \& Learn \cite{misra2016shuffle} \textcolor[rgb]{0,0,1}{[ECCV, 2016]} &50.2 & 18.1 \\
    VGAN \cite{vondrick2016generating} \textcolor[rgb]{0,0,1}{[NeulPS, 2016]}& 52.1 & - \\
    OPN \cite{lee2017unsupervised} \textcolor[rgb]{0,0,1}{[ICCV, 2017]} & 56.3 & 22.1 \\
    %Jiasaw & 51.5 & 22.5 \\
    Geometry \cite{gan2018geometry} \textcolor[rgb]{0,0,1}{[CVPR, 2018]} & 55.1 & 23.3\\
    ST Puzzles \cite{kim2019self} \textcolor[rgb]{0,0,1}{[AAAI, 2019]} & 60.6 & 28.3\\
    Clip Order \cite{xu2019self} \textcolor[rgb]{0,0,1}{[CVPR, 2019]} & 65.6& 28.4\\
    Scratch & 60.5 & 21.2 \\
    ImageNet & 67.1 & 28.5 \\
    \hline
    STCR & 64.2 & 30.5 \\
    %STCR (Kinetics) & & \\
    STCR (64f) & 71.2& 38.4\\
    \hline
    Shuffle \& Learn \cite{misra2016shuffle}+STCR & 67.3 & 33.4 \\
    Clip Order \cite{xu2019self}+STCR &70.4 & 34.8 \\
    %+ Video Clip.
    Kinetics & 96.8 & 74.5\\
    \hline
    \end{tabular}
    }
    \caption{Comparing our STCR to pervious method in the literature on UCF101 and HMDB51. All methods use the same standard C3D architecture and the accuracies are averaged over three splits. Here STCR (64f) means we finetune STCR with 64 frame input, while the others are 16.}
    \label{tab:sota_action_recognition_cmp}
\end{table}

We report the average classification accuracy over 3 splits and compares with other existing self-supervised methods in Table \ref{tab:sota_action_recognition_cmp}.
We also show the accuracy from finetuned models which are pretrained on larger supervised datasets such as ImageNet and Kinetics. 
We can make several observations:
\emph{i}. Compared to train from scratch, our network achieves 50\% improvement over Scratch and even higher than ImageNet pretrained model in HMDB51.
\emph{ii}. Our STCR leads to significantly improvement with more frames input. The analyse of video clip length is given in Supplementary Section \textcolor[rgb]{1,0,0}{5}.
\emph{iii}. Our network not only achieves prominent performance but also combines well with the others methods, which demonstrates the generality of our method.

\subsection{Visualization of Spatio-Temporal Transformation}
\label{sec:vis_tt}

In this experiment, we compare a I3D baseline train from scratch against self-supervised pretrain. And we train 2 baselines with different configurations of pretrain: (a). Train I3D on split 1 of HMDB51 from random initialization with Xavier \cite{glorot2010understanding}.
(b). First perform self-supervised pretrain with Temporal Augment and then finetune on split 1 of HMDB51.

\begin{figure}[t]
    \centering
    \includegraphics[width=\linewidth]{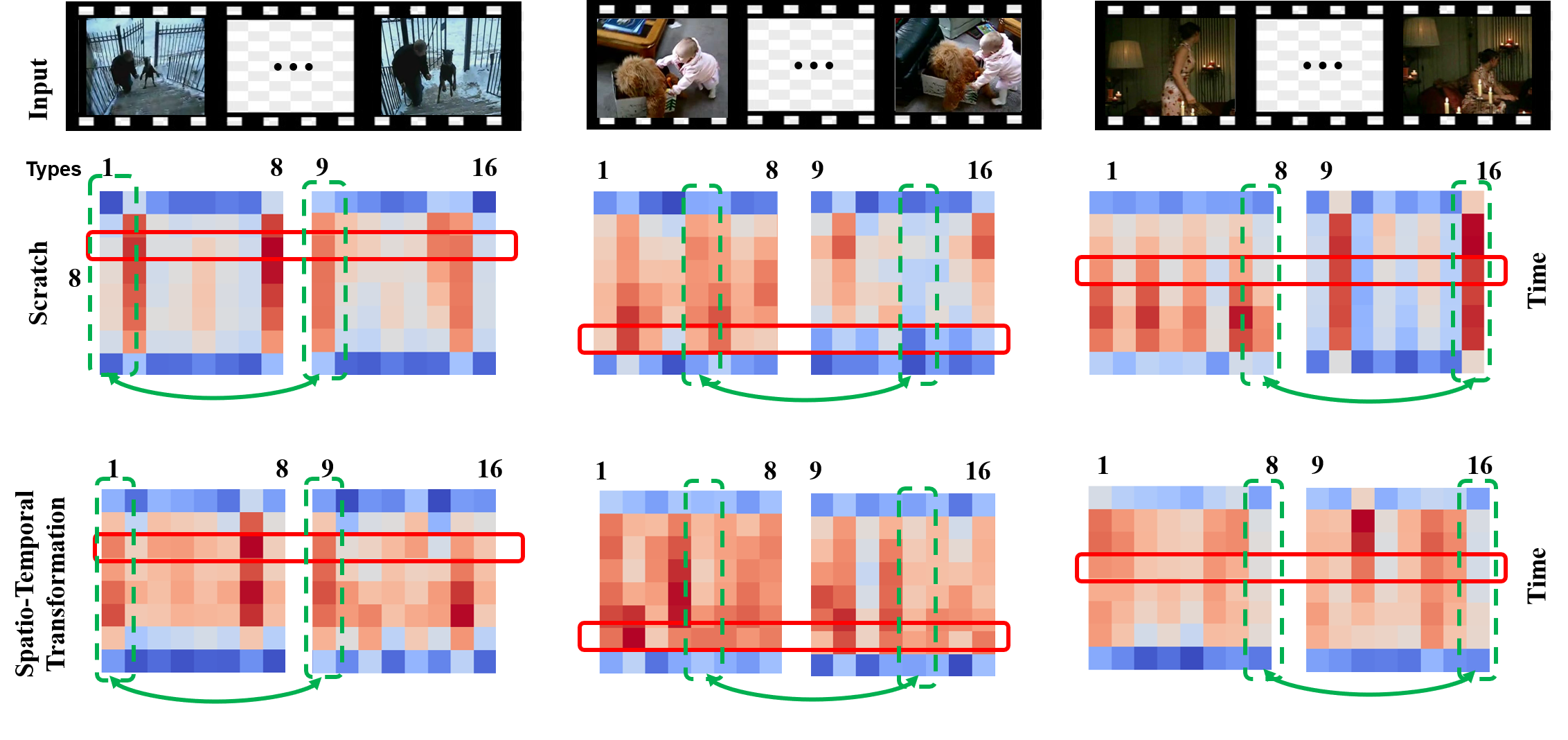}
    \caption{The network become more transformation consistency via Spatio-Temporal Transformation. These types of transformation correspond to Figure \ref{fig:stt_im}.}
    \label{fig:temporal_augment_visualization}
\end{figure}

After finish training, some feature maps generated by these two baselines are randomly selected and visualized.
More specifically, we select several samples and random crop a clip with shape $\mathbb{R}^{3 \times 64 \times 224 \times 224}$ for each sample.
$64$ is the number of frames, $3$ is the channel and $224$ is the width or height.
For each sample, $L$ kinds (16 in this paper) Spatio-Temporal Transformations are performed.
Then these samples are fed into the I3D.
When they feedforward into the layer \textit{Mixed\_5c}, the high-level feature map is $\mathbb{R}^{1024 \times 8 \times 7 \times 7}$. 
%We  .
For these high-level feature maps, we first perform inverse transformation, and then spatial global max pooling is performed first and average pooling along with all channels is followed. 
So that each transform sample is encoded into a description vector with shape $8$. 
%In order to .
We further stacking these description vectors horizontally as video-level representation, so that the overall matrix is $16 \times 8$. 
%The video-level representation from noise path should keep consistency with the clean one after inverse transformation. 

The result of this experiment is shown in Figure \ref{fig:temporal_augment_visualization}.
The only difference between the feature from the left half ($c \in [1, 8]$) and the right half ($c + 8$) are with or without adopt temporal flip in Temporal Transformation.
It is worth noting that the rotation invariance is barely encoded into the video-level representation under our STCR framework. As shown in Figure \ref{fig:temporal_augment_visualization} the feature maps are various in the same red boxes.
However, if the Spatio--Temporal Transformation plays a role as consistency constraint in STCR, then at least the feature map ($c$) from the left half of matrix should keep ``consistent'' with the right one ($c + 8$).
Compared the green boxes in Figure \ref{fig:temporal_augment_visualization}, it indicated that the features using Spatio-Temporal Transformation (third row) are more transformation ``consistent'' than the feature without self-supervised pre-train (second row).

% \subsection{Going deeper into Intra-Video Mixup}
\subsection{Insight Analysis of Intra-Video Mixup}

\label{sec:deeper_intra_sample_mixup}
%here give a image about which class improve and which class down
Besides the results of classification accuracy, this subsection further investigate how Intra-Video Mixup can boost the performance by visualizing representative videos and analyzing performance statistics of different categories.
% Beyond the overall classification accuracy, how Intra-Video Mixup boosting the performance? And in what cases exactly does it help? 
% To answer these questions, at first we visualize some representative videos, and then we analyse the category statistics. 

\begin{figure}[t]
    \centering
    \setlength{\belowcaptionskip}{1pt}
    \includegraphics[width=.9\linewidth]{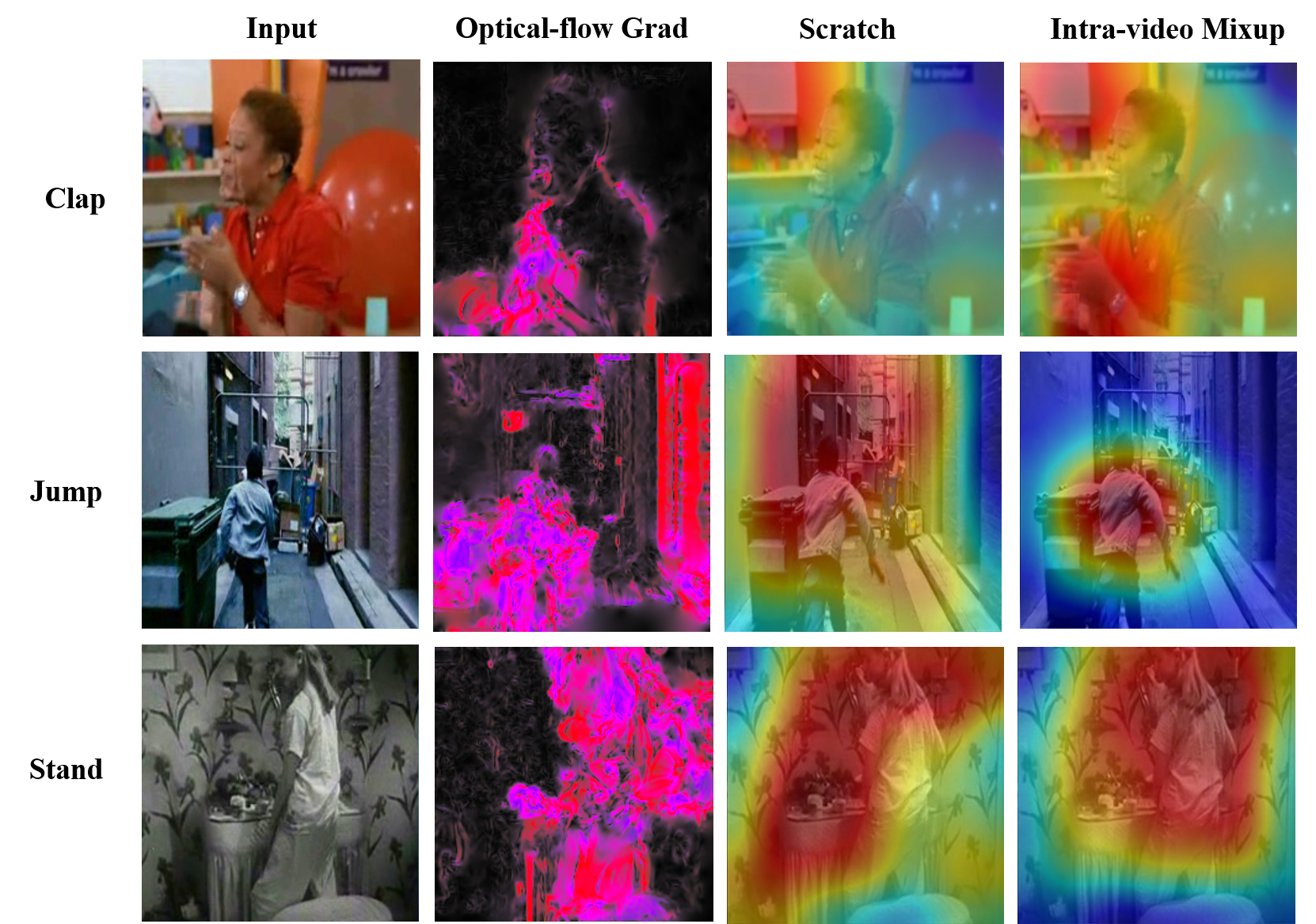}
    \caption{Comparison between random initialization (scratch) and self-supervised pre-train with our prosed Intra-Video Mixup. The model trained with Intra-Video Mixup care more about motion region. (Best viewed in color).}
    \label{fig:mixup_scratch_visualization}
\end{figure}

\textbf{Heatmap Visualization.}
Figure \ref{fig:mixup_scratch_visualization} visualizes the heatmaps of some samples in HMDB51 before and after self-supervised train, using I3D as baseline. 
% The only difference is the parameters of I3d is random initialization with Xavier \cite{glorot2010understanding} (the second row) or self-supervised via Intra-Video Mixup (the third row).
%Specially, we select some representative samples and each sample with shape $3 \times 64 \times 224 \times 224$ and then feed into the I3D baseline. 
%Then the high-level feature map size of each sample in \textit{Mixed\_5c} is $1024 \times  8 \times 7 \times 7$ and $1024$ is the number of channel. 
Specially, we select some representative samples with significant movement information and then each sample is encoded into $\mathbb{R}^{1024 \times  8 \times 7 \times 7}$ as in Section \ref{sec:vis_tt}.
For each channel we perform global max pooling along temporal dimension and enlarge this feature into input size, and then we average the feature map over all channels.
In Figure \ref{fig:mixup_scratch_visualization}, the second row (random initialization with Xavier \cite{glorot2010understanding}) and the third row (self-supervised via Intra-Video Mixup) show different heatmaps, which can observe that the network focus more on motion area after self-supervised pretrain.

\begin{figure}[t]
 \setlength{\belowcaptionskip}{5pt}
    \centering
    \includegraphics[width=1\linewidth]{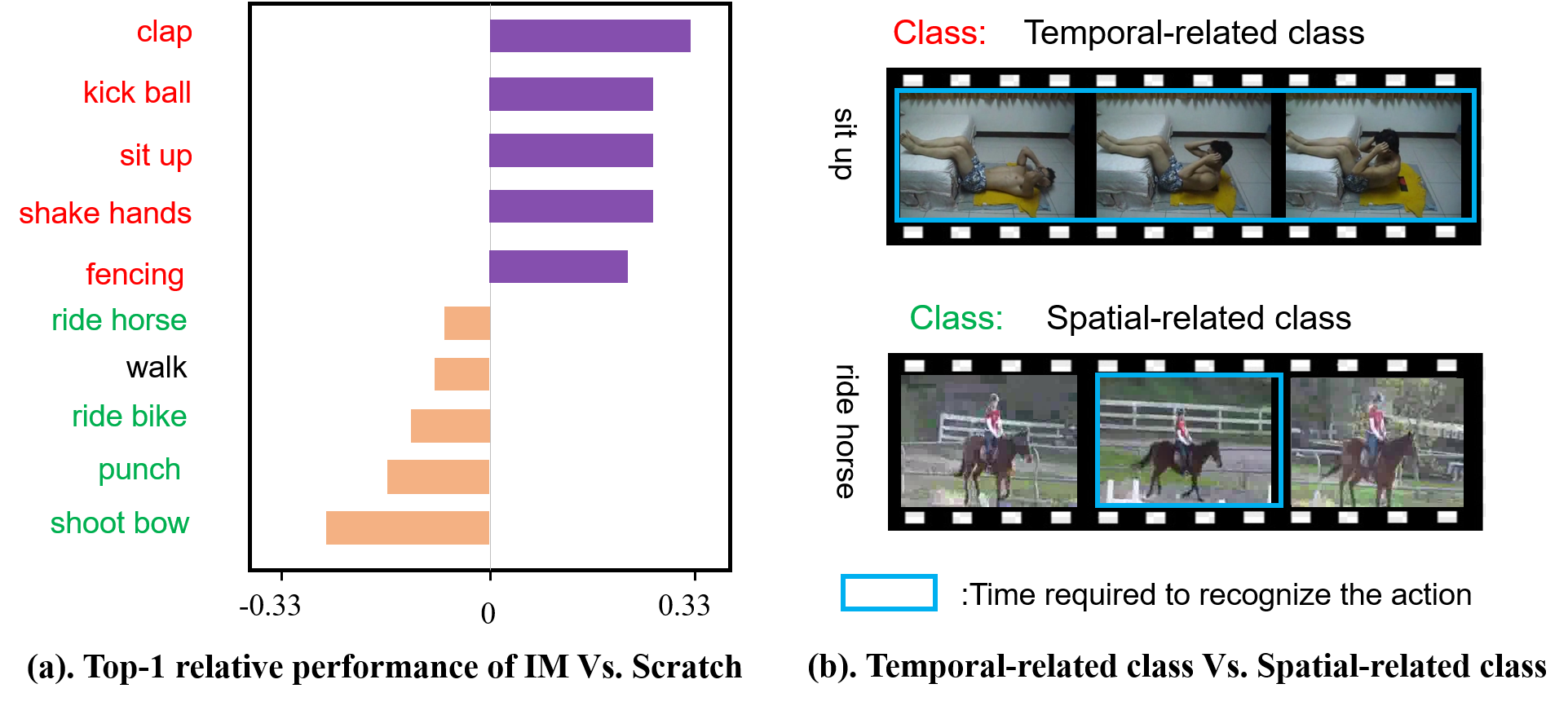}
    \caption{Intra-Video Mixup (IM) outperforms the Scratch when actions are temporal-related, in red. But when actions with prominent class-related visual features, Scratch performs better, in green. The overall performance of IM significantly exceeds Scratch baseline.}
    \label{fig:spatial_mixup_acc_diff}
\end{figure}

\textbf{Category statistics.}
Furthermore, in order to assess the relative performance of Intra-Video Mixup, we make a comparison between scratch and pre-train with Intra-Video Mixup.
The result are shown in Figure \ref{fig:spatial_mixup_acc_diff}, in which Intra-Video Mixup excels in temporal-related action categories.
The temporal-related action refers to the actions fully relying on motion information and can't be distinguished in one or a few frames. 
For example, the HMDB51 dataset has many action categories contain ball, like ``kick\_ball'', ``shoot\_ball'' and ``dribble''.
Among these ball-related actions, the model need to focus on motion pattern rater than spatial details to distinguish the specific class (``kick\_ball'') from the others (``shoot\_ball'' and ``dribble''). 
In contrast, Scratch excels in the cases where the action is spatial-related.
For example, ``shoot bow" and ``ride bike" can be distinguish in a frame with ``bow" and ``bike". 
It makes sense when considering the Intra-Video Mixup as a ``copy'' of static image and the ``bow" or ``bike" might be viewed as noise.
The above observations indicated that equipping Intra-Video Mixup is able to learn more temporal abstractions and suppress static image features, which is crucial for temporal-related actions.

\begin{figure}[t]
 \setlength{\belowcaptionskip}{5pt}
    \centering
    \includegraphics[width=\linewidth]{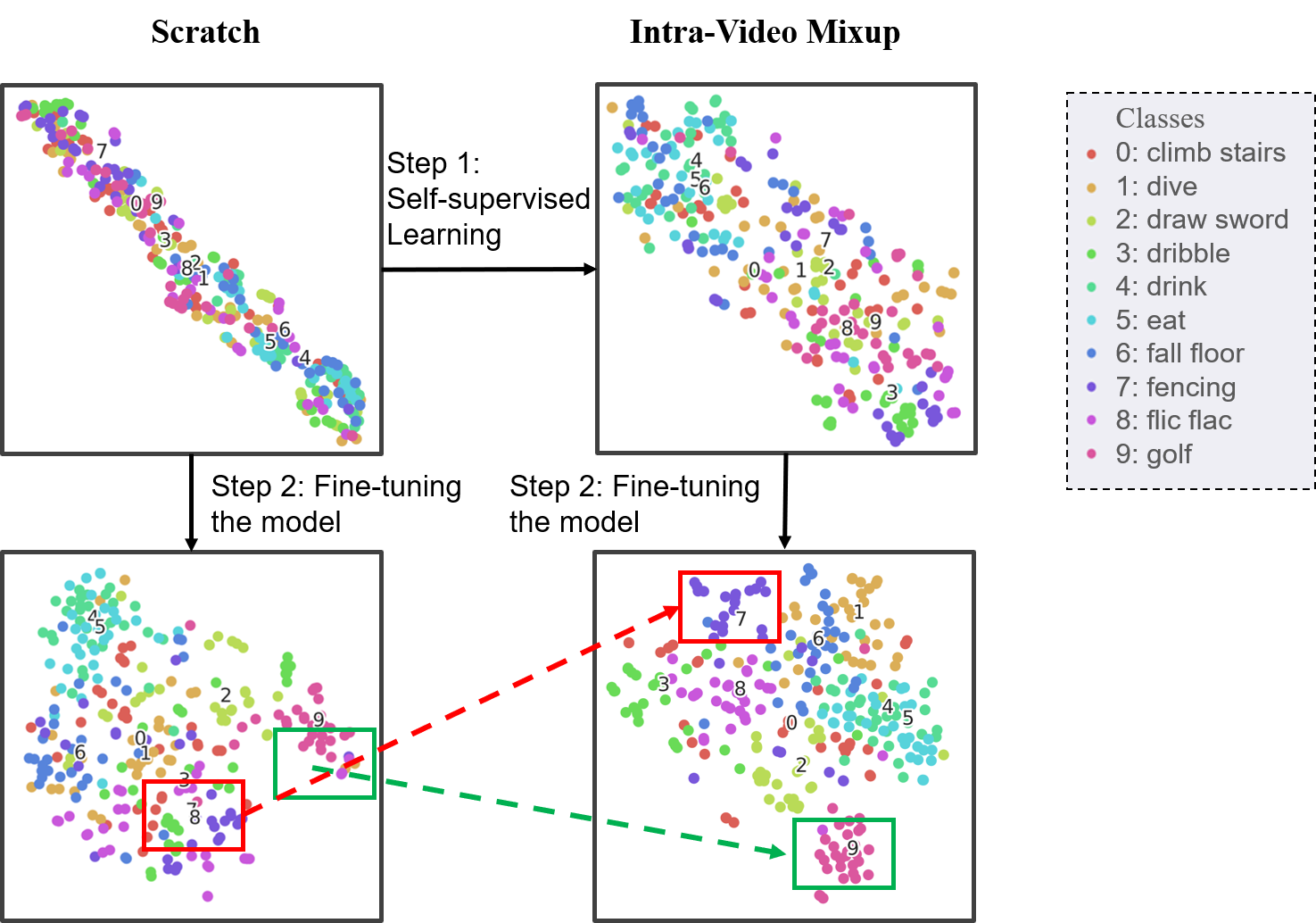}
    \caption{The visualization of t-SNE (Intra-video Mixup vs. Scratch). We random select 10 classes from HMDB51. The feature in embedding space become more diversity with Intra-Video Mixup, which suggests that more easily separable features are learned.}
    \label{fig:t-sne}
\end{figure}

We further show the t-SNE visualization result of Intra-Video Mixup in Figure \ref{fig:t-sne}. What is obvious is that after the self-supervised learning with Intra-Video Mixup, these samples in embedding space become more diversity (first row). We also observe that Intra-Video Mixup has better ``clusters'' (red and green box) as we fine-tune on downstream action recognition task.
\section{Conclusion}
In this paper, we propose a novel Spatio-Temporal Consistency Regularization (STCR) method for self-supervised learning. 
% constrains consistency bewteen the augmented data by Spatio-Temporal Transformation and Intra-Video Mixup
The proposed method minimizes the variations in different passes of a sample through the network caused by Spatio-Temporal Transformation and Intra-Video Mixup. 
The proposed method is evaluated by using different CNN backbones on two benchmark datasets. 
Experimental results show that the proposed STCR outperforms existing methods for action recognition without labelled data for pretraining.

Our future work will study how to make use of other advanced consistency learning methods in the self-supervised setting. On the other hand, besides action recognition, we will further develop self-supervised learning method based on the proposed STCR for other video applications like spatio-temporal action localization.

% most modern works on video self-supervised learning algorithms are evaluated on video action recognition benchmarks.
% We are also interested in exploring the effectiveness of STCR in other domains.

\bibliographystyle{unsrt}  
\bibliography{main}

\begin{thebibliography}{10}

\bibitem{hara2018can}
Kensho Hara, Hirokatsu Kataoka, and Yutaka Satoh.
\newblock Can spatiotemporal 3d cnns retrace the history of 2d cnns and
  imagenet?
\newblock In {\em CVPR}, pages 6546--6555, 2018.

\bibitem{zhao2017temporal}
Yue Zhao, Yuanjun Xiong, Limin Wang, Zhirong Wu, Xiaoou Tang, and Dahua Lin.
\newblock Temporal action detection with structured segment networks.
\newblock In {\em ICCV}, pages 2914--2923, 2017.

\bibitem{weinzaepfel2015learning}
Philippe Weinzaepfel, Zaid Harchaoui, and Cordelia Schmid.
\newblock Learning to track for spatio-temporal action localization.
\newblock In {\em ICCV}, pages 3164--3172, 2015.

\bibitem{zhai2019s}
Xiaohua Zhai, Avital Oliver, Alexander Kolesnikov, and Lucas Beyer.
\newblock S4l: Self-supervised semi-supervised learning.
\newblock {\em ICCV}, 2019.

\bibitem{noroozi2016unsupervised}
Mehdi Noroozi and Paolo Favaro.
\newblock Unsupervised learning of visual representations by solving jigsaw
  puzzles.
\newblock In {\em ECCV}, pages 69--84. Springer, 2016.

\bibitem{gidaris2018unsupervised}
Spyros Gidaris, Praveer Singh, and Nikos Komodakis.
\newblock Unsupervised representation learning by predicting image rotations.
\newblock In {\em ICLR}, 2018.

\bibitem{wei2018learning}
Donglai Wei, Joseph~J Lim, Andrew Zisserman, and William~T Freeman.
\newblock Learning and using the arrow of time.
\newblock In {\em CVPR}, pages 8052--8060, 2018.

\bibitem{xu2019self}
Dejing Xu, Jun Xiao, Zhou Zhao, Jian Shao, Di~Xie, and Yueting Zhuang.
\newblock Self-supervised spatiotemporal learning via video clip order
  prediction.
\newblock In {\em CVPR}, pages 10334--10343, 2019.

\bibitem{soomro2012ucf101}
Khurram Soomro, Amir~Roshan Zamir, and Mubarak Shah.
\newblock Ucf101: A dataset of 101 human actions classes from videos in the
  wild.
\newblock {\em arXiv preprint arXiv:1212.0402}, 2012.

\bibitem{kuehne2013hmdb51}
Hilde Kuehne, Hueihan Jhuang, Rainer Stiefelhagen, and Thomas Serre.
\newblock Hmdb51: A large video database for human motion recognition.
\newblock In {\em High Performance Computing in Science and Engineering ?12},
  pages 571--582. Springer, 2013.

\bibitem{ye2019unsupervised}
Mang Ye, Xu~Zhang, Pong~C Yuen, and Shih-Fu Chang.
\newblock Unsupervised embedding learning via invariant and spreading instance
  feature.
\newblock In {\em CVPR}, pages 6210--6219, 2019.

\bibitem{kim2019self}
Dahun Kim, Donghyeon Cho, and In~So Kweon.
\newblock Self-supervised video representation learning with space-time cubic
  puzzles.
\newblock In {\em AAAI}, volume~33, pages 8545--8552, 2019.

\bibitem{sajjadi2016regularization}
Mehdi Sajjadi, Mehran Javanmardi, and Tolga Tasdizen.
\newblock Regularization with stochastic transformations and perturbations for
  deep semi-supervised learning.
\newblock In {\em NeurlPS}, pages 1163--1171, 2016.

\bibitem{tarvainen2017mean}
Antti Tarvainen and Harri Valpola.
\newblock Mean teachers are better role models: Weight-averaged consistency
  targets improve semi-supervised deep learning results.
\newblock In {\em NeurlPS}, pages 1195--1204, 2017.

\bibitem{berthelot2019mixmatch}
David Berthelot, Nicholas Carlini, Ian Goodfellow, Nicolas Papernot, Avital
  Oliver, and Colin Raffel.
\newblock Mixmatch: A holistic approach to semi-supervised learning.
\newblock {\em NeurlPS}, 2019.

\bibitem{xie2019unsupervised}
Qizhe Xie, Zihang Dai, Eduard Hovy, Minh-Thang Luong, and Quoc~V Le.
\newblock Unsupervised data augmentation for consistency training.
\newblock {\em arXiv preprint arXiv:1904.12848}, 2019.

\bibitem{Guo_2019_CVPR}
Hao Guo, Kang Zheng, Xiaochuan Fan, Hongkai Yu, and Song Wang.
\newblock Visual attention consistency under image transforms for multi-label
  image classification.
\newblock In {\em CVPR}, June 2019.

\bibitem{zhou2016learning}
Bolei Zhou, Aditya Khosla, Agata Lapedriza, Aude Oliva, and Antonio Torralba.
\newblock Learning deep features for discriminative localization.
\newblock In {\em CVPR}, pages 2921--2929, 2016.

\bibitem{krizhevsky2012imagenet}
Alex Krizhevsky, Ilya Sutskever, and Geoffrey~E Hinton.
\newblock Imagenet classification with deep convolutional neural networks.
\newblock In {\em NeurlPS}, pages 1097--1105, 2012.

\bibitem{zhang2018mixup}
Yann N.~Dauphin Hongyi~Zhang, Moustapha~Cisse and David Lopez-Paz.
\newblock mixup: Beyond empirical risk minimization.
\newblock {\em ICLR}, 2018.

\bibitem{paszke2017automatic}
Adam Paszke, Sam Gross, Soumith Chintala, Gregory Chanan, Edward Yang, Zachary
  DeVito, Zeming Lin, Alban Desmaison, Luca Antiga, and Adam Lerer.
\newblock Automatic differentiation in pytorch.
\newblock 2017.

\bibitem{tran2015learning}
Du~Tran, Lubomir Bourdev, Rob Fergus, Lorenzo Torresani, and Manohar Paluri.
\newblock Learning spatiotemporal features with 3d convolutional networks.
\newblock In {\em ICCV}, pages 4489--4497, 2015.

\bibitem{tran2018closer}
Du~Tran, Heng Wang, Lorenzo Torresani, Jamie Ray, Yann LeCun, and Manohar
  Paluri.
\newblock A closer look at spatiotemporal convolutions for action recognition.
\newblock In {\em CVPR}, pages 6450--6459, 2018.

\bibitem{carreira2017quo}
Joao Carreira and Andrew Zisserman.
\newblock Quo vadis, action recognition? a new model and the kinetics dataset.
\newblock In {\em CVPR}, pages 4724--4733. IEEE, 2017.

\bibitem{misra2016shuffle}
Ishan Misra, C~Lawrence Zitnick, and Martial Hebert.
\newblock Shuffle and learn: unsupervised learning using temporal order
  verification.
\newblock In {\em ECCV}, pages 527--544. Springer, 2016.

\bibitem{vondrick2016generating}
Carl Vondrick, Hamed Pirsiavash, and Antonio Torralba.
\newblock Generating videos with scene dynamics.
\newblock In {\em NeurlPS}, pages 613--621, 2016.

\bibitem{lee2017unsupervised}
Hsin-Ying Lee, Jia-Bin Huang, Maneesh Singh, and Ming-Hsuan Yang.
\newblock Unsupervised representation learning by sorting sequences.
\newblock In {\em ICCV}, pages 667--676, 2017.

\bibitem{gan2018geometry}
Chuang Gan, Boqing Gong, Kun Liu, Hao Su, and Leonidas~J Guibas.
\newblock Geometry guided convolutional neural networks for self-supervised
  video representation learning.
\newblock In {\em CVPR}, pages 5589--5597, 2018.

\bibitem{glorot2010understanding}
Xavier Glorot and Yoshua Bengio.
\newblock Understanding the difficulty of training deep feedforward neural
  networks.
\newblock In {\em AISTATS}, pages 249--256, 2010.

\end{thebibliography}

\end{document}